\documentclass[letterpaper]{article}

\usepackage{aaai}
\usepackage{times}
\usepackage{bbm} 
\usepackage{mathtools}
\usepackage{tikz} 
\usepackage{pgfplots}
\usepackage{amssymb}
\usepackage{amsmath}
\usepackage{amsfonts}
\usepackage{fullpage}
\usepackage{multirow}
\usepackage{tikz}
\usetikzlibrary{shapes,arrows}
\usepackage{bm}

\newcommand{\vnorm}[1]{\left|\left|#1\right|\right|}
\newcommand{\indic}[1]{\mathbbm{1}[#1]}

\begin{document}
\title{Supersparse Linear Integer Models for Predictive Scoring Systems}
\author{Berk Ustun, Stefano Trac\`a, Cynthia Rudin\\
Massachusetts Institute of Technology\\
77 Massachusetts Avenue, Cambridge, MA 02139, USA\\
}
\maketitle
%\begin{abstract}
%\begin{quote}
%Abstract to be written last.
%\end{quote}
%\end{abstract}

\noindent 
Scoring systems are classification models that make predictions using a sparse linear combination of variables with integer coefficients. Such systems are frequently used in medicine because they are interpretable; that is, they only require users to add, subtract and multiply a few meaningful numbers in order to make a prediction. See, for instance, these commonly used scoring systems: \cite{gage2001validation,le1984simplified,le1993new,knaus1985apache}. Scoring systems strike a delicate balance between accuracy and interpretability that is difficult to replicate with existing machine learning algorithms. 

Current linear methods such as the lasso, elastic net and LARS are not designed to create scoring systems, since regularization is primarily used to improve accuracy as opposed to sparsity and interpretability \cite{tibshirani1996regression,zou2005regularization,efron2004least}. These methods can produce very sparse models through heavy regularization or feature selection methods \cite{guyon2003introduction}; however, feature selection often relies on greedy optimization and cannot guarantee an optimal balance between sparsity and accuracy. Moreover, the interpretability of scoring systems requires integer coefficients, which these methods do not produce. Existing approaches to interpretable modeling include decision trees and lists \cite{ruping2006learning,quinlan1986induction,rivest1987learning,LethamRuMcMa12}.

We introduce a formal approach for creating scoring systems, called Supersparse Linear Integer Models (SLIM). SLIM produces scoring systems that are accurate and interpretable using a mixed-integer program (MIP) whose objective penalizes the training error, $L_0$-norm and $L_1$-norm of its coefficients. SLIM can create scoring systems for datasets with thousands of training examples and tens to hundreds of features -  larger than the sizes of most studies in medicine, where scoring systems are often used.

\section{Formulation} 

Given a dataset with $N$ examples and $P$ features, SLIM produces classifier $\hat{y}=\mathrm{sign}(\mathbf{x}^T{\bm\lambda})$ where $\mathbf{x}\in\mathbb{R}^P$ is a vector of features, $\hat{y}\in\{-1,1\},$ are predicted labels and $\bm{\lambda}\in \mathbb{Z}^P$ is a vector of coefficients. The optimization for SLIM on $N$ training examples is: 
\begin{align*}\centering
\min_{\bm{\lambda}} \, \frac{1}{N} \sum_{i=1}^{N}\indic{y_i \mathbf{x}_i^T \bm{\lambda} \leq 0} + C_0\vnorm{\bm{\lambda}}_0 + C_1\vnorm{\bm{\lambda}}_1 
\end{align*}
where $C_0$ and $C_1$ are penalties associated with the $L_0$-norm and $L_1$-norm of $\bm{\lambda}$. $C_0$ is the primary parameter, inducing a sparse set of coefficients, and $C_1$ is the secondary parameter ensuring that the coefficients are as small (and interpretable) as possible. Note $C_1$ does not take away from the sparsity of the solution, but promotes the smallest coefficients among equally sparse models. The objective is minimized by a MIP with $N + 3P$ variables and $2N + 4P$ constraints:
\begin{alignat*}{3}\centering
&\min_{\bm{\alpha},\bm{\beta},\bm{\gamma},\bm{\lambda}}\,\frac{1}{N}\sum_{i=1}^{N}\alpha_i \tiny{+} C_0\sum_{j=1}^{P}\beta_j + C_1\sum_{j=1}^{P}{\gamma_j}&\\
& -M\alpha_i + \epsilon \leq y_i \mathbf{x}_i^T \bm{\lambda} \leq M(1-\alpha_i) + \epsilon & i= 1\ldots N \\ 
                & \qquad -\Lambda \beta_j  \leq \quad \lambda_j \quad \leq \Lambda \beta_j & j= 1\ldots P \\ 
                & \qquad \,\,\,\, -\gamma_j \leq \quad \lambda_j \quad \leq \gamma_j & j= 1\ldots P \\ 
                & \qquad\qquad\qquad\bm{\lambda} \in \mathcal{L} & \\
                & \qquad\qquad\quad \,\,\,\,\alpha_i \in \{0,1\} & i =1\ldots N \\
                & \qquad \quad \,\beta_j  \in \{0,1\}, \gamma_j \in \mathbb{R}_+\; &  j= 1\ldots P \\ 
\end{alignat*}
where $\alpha_i = \indic{y_i\neq \hat{y}_i}$, $\beta_j = \indic{\lambda_j\neq 0}$, and $\gamma_j = |\lambda_j|$. All feasible $\bm{\lambda}$ belong to $\{\bm{\lambda} \in \mathcal{L}: |\lambda_j| \leq \Lambda \;\forall j\}$. By default, we set $\Lambda = 100$ and $\mathcal{L} = \mathbb{Z}^P$, although we often further restrict the coefficients to have only one significant digit. Lastly, $\epsilon$ and $M$ are scalars used in if-then constraints; we set $\epsilon = 0.1$, $M =  \Lambda \cdot
\max_{i,j} |x_{ij}|$.

Computational factors affect the accuracy and sparsity of SLIM scoring systems: current MIP solvers can train sparse and accurate scoring systems for datasets with $N \approx 10000$ and $P \approx 100$.
%; in a future publication, we will demonstrate that such models can be produced for larger datasets using Tabu Search.

\section{Theoretical Bound} 

We can bound the true risk of a SLIM scoring system, $R^{\text{true}}(f)=\mathbb{E}_{X,Y}\mathbbm{1}[f(X) \neq Y] $, by its empirical risk, $R^{\text{emp}}(f)=\frac{1}{N}\sum_{i=1}^N \mathbbm{1}[f(\mathbf{x}_i) \neq y_i] $, as follows: \\
For every $ \delta > 0,$ and every $ f \in \{f_1, \cdots, f_K \}$,
\begin{align*} 
R^{\text{true}}(f) \leq R^{\text{emp}}(f) + \sqrt{\frac{\log(K) - \log(\delta)}{2N}} ,
\end{align*}
with probability at least $1-\delta$. Here, $f(\mathbf{x}_i)=\text{sign}(\mathbf{x}_i^T\bm{\lambda})$ and $\bm{\lambda} \in \{\bm{\lambda} \in \mathcal{L}: |\lambda_j| \leq \Lambda \}$. Since $\mathcal{L} \subseteq \mathbb{Z}^P$, it can be shown that $\log(K) 
%\leq \log(|\mathcal{L}|) 
\leq P \log(2\Lambda + 1)$.

\section{Experimental Results}

We compare the accuracy and sparsity of SLIM scoring systems to classification models produced by C5.0, CART, Logistic Regression (LR), Elastic Net (EN), Random Forests (RF) and Support Vector Machines (SVM) in Table \ref{comparison_table}. Our comparison includes the breastcancer, haberman, internetad, mammo, spambase and tictactoe datasets from the UCI Machine Learning Repository, as well as the violent-crime dataset, which is derived from a study of crime among young people raised in out-of-home care, made available by the US Department of Justice Statistics \cite{cusick2010crime}. 

We report the mean 5-fold cross-validation (CV) test error as a measure of accuracy (top), and the median 5-fold CV model size as a measure of sparsity (bottom). Model size reflects the number of coefficients for LR, EN and SLIM, and the number of leaves for C5.0 and CART; we omit this statistic for RF and SVM as sparsity does not affect the interpretability of these methods. SLIM models were trained for 1 hour using the CPLEX 12.5 API for MATLAB; all other models were trained for default times using packages in R 2.15. We set the free parameters for most methods to values that minimized the 5-fold CV test error; for EN, we set the $L_1$ penalty to the value that produced the sparsest model on the $L_1$ regularization path, within 1-SE of the $L_1$ penalty that minimized the 5-fold CV test error. 

\begin{table}[htbp]
  \centering
    \resizebox{8cm}{!} {
    \begin{tabular}{|c|c|c|c|c|c|c|c|c|c|}
    \hline
    Dataset & $N$ & $P$ & C5.0 & CART & LR & EN & RF & SVM & SLIM \\
    \hline
    \multirow{2}[2]{*}{breastcancer} & \multirow{2}[2]{*}{683} & \multirow{2}[2]{*}{10} & 5.3\% & 5.9\% & 3.7\% & 3.5\% & 2.7\% & 2.9\% & 3.7\% \\
      &   &   & 8 & 4 & 9 & 10 & - & - & 3 \\
    \hline
    \multirow{2}[2]{*}{haberman} & \multirow{2}[2]{*}{306} & \multirow{2}[2]{*}{4} & 27.8\% & 26.8\% & 26.5\% & 26.5\% & 28.1\% & 26.2\% & 23.2\% \\
      &   &   & 3 & 6 & 3 & 4 & - & - & 3 \\
    \hline
    \multirow{2}[2]{*}{internetad} & \multirow{2}[2]{*}{2359} & \multirow{2}[2]{*}{1431} & 3.9\% & 4.5\% & 8.5\% & 3.1\% & 2.5\% & 3.7\% & 3.6\% \\
      &   &   & 10 & 7 & 616 & 473 & - & - & 14 \\
    \hline
    \multirow{2}[2]{*}{mammo} & \multirow{2}[2]{*}{961} & \multirow{2}[2]{*}{6} & 18.4\% & 17.5\% & 29.4\% & 32.4\% & 18.0\% & 17.2\% & 17.2\% \\
      &   &   & 5 & 4 & 5 & 4 & - & - & 4 \\
    \hline
    \multirow{2}[2]{*}{spambase} & \multirow{2}[2]{*}{4601} & \multirow{2}[2]{*}{58} & 7.7\% & 10.6\% & 7.3\% & 7.6\% & 4.8\% & 6.5\% & 7.4\% \\
      &   &   & 63 & 7 & 57 & 52 & - & - & 18 \\
    \hline
    \multirow{2}[2]{*}{tictactoe} & \multirow{2}[2]{*}{958} & \multirow{2}[2]{*}{28} & 7.5\% & 11.7\% & 2.7\% & 1.7\% & 1.6\% & 0.7\% & 3.3\% \\
      &   &   & 39 & 21 & 18 & 19 & - & - & 18 \\
    \hline
    \multirow{2}[2]{*}{violentcrime} & \multirow{2}[2]{*}{558} & \multirow{2}[2]{*}{108} & 25.1\% & 24.9\% & 22.2\% & 19.5\% & 21.3\%	& 19.5\%  &  20.1\% \\
      &   &   & 27 &  10 & 57 & 108 & - & - & 9 \\
          \hline
    \end{tabular}%
    }
    \caption{Accuracy and sparsity of all methods.}
    \label{comparison_table}%
\end{table}

Our results suggest that these methods produce classification models that are comparable in terms of accuracy but vary \emph{dramatically} in terms of sparsity. In particular, SLIM consistently produces scoring systems that are both accurate and sparse. Furthermore, as shown in the following two demonstrations, SLIM also produces scoring systems that are highly interpretable.

\subsection{A Scoring System to Detect Breast Cancer}

When applied to the breastcancer dataset \cite{MangaWo90}, SLIM produces a scoring system to predict whether a tumor is malignant (Class $ = +1$) using only 3 cell-related features. Here, coefficients are restricted to $\mathcal{L} = \{0,\pm 1, \pm 5, \pm 10, \pm 50, \pm 100, \pm 500 \}^{10}$: 
{\small
\begin{alignat*}{2}\centering
\text{Score} = &\, \text{ClumpThickness} + \,\text{UniformityOfCellSize} \\  
      &   \,\,+ \text{BareNuclei} - 10 \\
\text{Predicted Class} = &\text{sign}(\text{Score}).
\end{alignat*}% 
}
\noindent This score would be easy for doctors to compute when analyzing patient scans.

\subsection{A Scoring System to Predict Violent Crime}
 
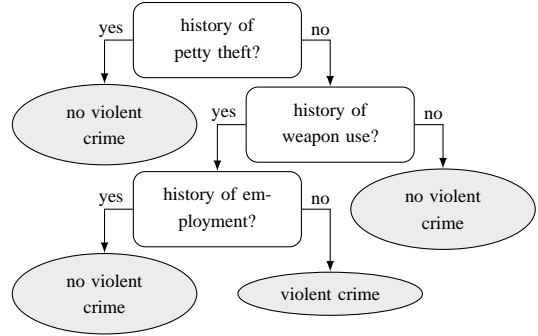
\begin{figure}
\tikzset{
    every node/.style={
        font=\scriptsize
    },
    decision/.style={
        shape=rectangle,
        minimum height=1cm,
        text width=2cm,
        text centered,
        rounded corners=1ex,
        draw,
        label={[yshift=0.125cm]left:yes},
        label={[yshift=0.125cm]right:no},
    },
    outcome/.style={
        shape=ellipse,
        fill=gray!15,
        draw,
        text width=1.5cm,
        text centered
    },
    decision tree/.style={
        edge from parent path={[-latex] (\tikzparentnode) -| (\tikzchildnode)},
        sibling distance=3cm,
        level distance=1.125cm
    }
}

\begin{tikzpicture}
\node [decision] { history of petty theft? }
    [decision tree]
    child { node [outcome] { no violent crime } }
    child { node [decision] { history of weapon use?} 
        child { node [decision] { history of employment? } 
            child { node [outcome] { no violent crime } }
            child { node [outcome] { violent crime } }
        }
        child { node [outcome] { no violent crime } }
    };

\end{tikzpicture}
\caption{Decision tree induced by a SLIM model. \label{tree_model}}
\end{figure}

When applied to the violentcrime dataset, SLIM produces the following scoring system to predict whether a young person raised in out-of-home care will commit a violent crime (Class $= + 1$) using 3 features related to their background and criminal record: 
{\small
\begin{alignat*}{2}\centering
\textrm{Score} = &-10\, \textrm{PettyTheft} +9\,  \textrm{WeaponUse}  \\
      &-9\,  \textrm{Employment} -1\\
\text{Predicted Class} = &\text{sign}(\text{Score}).
\end{alignat*}%
}
\noindent The three features are indicators for past history of petty theft, indicator of past weapon use, and whether the person has ever been employed.
In this case, we had restricted coefficients to the set $\mathcal{L} = \{0,\pm 1,\pm 9,\pm 10\}^{108}$. This model can also be visualized as a decision tree since it uses only discrete features, as shown in Figure \ref{tree_model}.\\

\section{Conclusion}
We introduced SLIM as a tool to create data-driven scoring systems for binary classification. 
%In addition, we have derived theoretical bounds on the true risk of SLIM scoring systems, and
SLIM's models tend to be accurate since they are optimized, but also highly interpretable, as they are built from a small number of non-zero terms with integer coefficients.
%Our results indicate that SLIM is able to produce models that are comparable to the state-of-the-art in accuracy, but at the same time, highly interpretable, in that they possess a sparse number of nonzero, integer coefficients.

\newpage

\bibliographystyle{aaai}
\bibliography{slim_aaai}

\end{document}